%% file: main.tex
\documentclass[10pt,twocolumn,letterpaper]{article}

\usepackage{iccv}              


\definecolor{iccvblue}{rgb}{0.21,0.49,0.74}
\usepackage[pagebackref,breaklinks,colorlinks,allcolors=iccvblue]{hyperref}

\usepackage{color}
\usepackage{amsmath}
\usepackage{bm}

\usepackage{graphicx}
\usepackage{colortbl}
\usepackage{xcolor}
\usepackage{multirow}

\usepackage{caption}
\usepackage{algorithm}
\usepackage{algpseudocode}
\usepackage{tikz}
\usetikzlibrary{arrows}
\usepackage{tabulary}
\usepackage{amssymb}
\usepackage{booktabs}
\usepackage{booktabs}

\def\paperID{2982} 
\def\confName{ICCV}
\def\confYear{2025}

\title{HEMGS: A Hybrid Entropy Model for 3D Gaussian Splatting Data Compression}
\author{Lei Liu$^1$, \quad Zhenghao Chen$^2$, \quad Wei Jiang$^3$, \quad Wei Wang$^3$, \quad Dong Xu$^1$ \\
$^1$ The University of Hong Kong \qquad  $^2$ The University of Newcastle \qquad $^3$ Futurewei Technologies Inc. \\
{\tt\small \{liulei95, dongxu\}@hku.hk, zhenghao.chen@newcastle.edu.au}
}

\begin{document}
\maketitle
\input{sec/0_abstract}

\input{sec/1_intro}
\input{sec/2_relatedwork}
\input{sec/3_methodology}
\input{sec/4_experiment}
{
    \small
    \bibliographystyle{ieeenat_fullname}
    \bibliography{main}
}

\end{document}

%% file: sec/0_abstract.tex
\begin{abstract}
%
In this work, we propose a novel compression framework for 3D Gaussian Splatting (3DGS) data. Building on anchor-based 3DGS methodologies, our approach compresses all attributes within each anchor by introducing a novel Hybrid Entropy Model for 3D Gaussian Splatting (HEMGS) to achieve hybrid lossy-lossless compression. It consists of three main components: a variable-rate predictor, a hyperprior network, and an autoregressive network. First, unlike previous methods that adopt multiple models to achieve multi-rate lossy compression, thereby increasing training overhead, our variable-rate predictor enables variable-rate compression with a single model and a hyperparameter $\lambda$ by producing a learned Quantization Step feature for versatile lossy compression.
Second, to improve lossless compression, the hyperprior network captures both scene-agnostic and scene-specific features to generate a prior feature, while the autoregressive network employs an adaptive context selection algorithm with flexible receptive fields to produce a contextual feature. By integrating these two features, HEMGS can accurately estimate the distribution of the current coding element within each attribute, enabling improved entropy coding and reduced storage.
We integrate HEMGS into a compression framework, and experimental results on four benchmarks indicate that HEMGS achieves about a 40\% average reduction in size while maintaining rendering quality over baseline methods and achieving state-of-the-art compression results.

\end{abstract}



%% file: sec/1_intro.tex
\section{Introduction}

In recent years, novel view synthesis has made significant progress in 3D scene representation.
Neural Radiance Field (NeRF)~\cite{mildenhall2021nerf} has appeared to represent 3D scenes as high rendering quality but suffers from slow rendering speed, limiting its practicality.
Recently, 3D Gaussian Splatting (3DGS)~\cite{kerbl20233d} has emerged as an efficient alternative, leveraging learnable Gaussians for explicit 3D modeling. With fast training and rendering, 3DGS has been widely adopted for novel view synthesis.
However, 3DGS requires substantial 3D Gaussians and associated attributes for effectively rendering scenes, which will bring significant challenges for storage or transmission.

To address this problem, early compression methods leveraged auxiliary compact data structures such as codebooks~\cite{fan2023lightgaussian,lee2024compact,navaneet2023compact3d,niedermayr2024compressed} and anchors~\cite{lu2024scaffold} to reduce storage. Although these compact structures can exploit the redundancy in raw 3DGS data, two major challenges persist. First, redundancy within the structures (\eg, anchors) is not completely eliminated. Second, these methods compress 3DGS into a single storage format, significantly limiting their flexibility and applicability in real-world scenarios that require varying storage options due to different conditions (\eg, network bandwidth).

Inspired by the success of neural image compression (NIC)~\cite{balle2018variational,minnen2018joint,cheng2020learned,dvc,fvc,li2021dcvc}, neural entropy models have been introduced for compressing 3DGS data~\cite{chen2025hac,wang2024contextgs} to tackle these challenges. These methods achieve multi-rate compression by applying rate-distortion (RD) optimizations with different rate-control hyperparameters, resulting in multiple models corresponding to multiple specific rates. Additionally, the neural entropy model is used to losslessly compress the quantized components at various compression rates, thereby reducing redundancy within the introduced structures (\eg, anchors).
However, these approaches still do not address the two issues effectively. First, for multi-rate lossy compression, using multiple separate models significantly increases training and storage overhead, especially when many compression rates are required. Second, these models typically do not fully exploit the neural entropy model for lossless compression, leaving room for further improvement.

We believe there is a more effective solution to address the two issues mentioned above and to achieve both variable-rate lossy compression and improved lossless compression. In this work, we propose a hybrid entropy model for 3D Gaussian Splatting data (HEMGS) that supports variable-rate lossy compression within a single model while also offering an enhanced neural entropy model for lossless compression.
Building on the anchor-based 3DGS methodology, our HEMGS compresses all attributes within a given anchor in a hybrid lossy-lossless fashion using quantization for lossy compression and a neural entropy coding strategy for lossless compression.
First, for variable-rate lossy compression of each attribute, we introduce a Variable-rate Predictor that accepts a hyperparameter $\lambda$ to adjust storage. This hyperparameter is fused with features derived from the previously compressed location and attribute information to generate a learned quantization step, which is then used in an adaptive quantization procedure~\cite{chen2025hac} for lossy compression at various storage rates. By enabling multi-rate compression with a single model, we reduce both the training overhead and model storage burden, making our lossy compression procedure more versatile for real-world applications with varying storage requirements.

On the other hand, to improve lossless compression of each attribute, we adopt a joint hyperprior and autoregressive model—an approach, inspired by NIC methods~\cite{minnen2018joint,cheng2020learned,5,6,9}, to predict the distribution of each element for entropy coding. Specifically, we introduce a hyperprior network that takes previously compressed location and attribute information as input. This network leverages a PointNet++ model pre-trained on a 3D domain to generate a scene-agnostic prior feature for better generalization, while employing MLPs to overfit a single 3DGS instance to produce a scene-specific prior feature. Consequently, unlike existing 3DGS data compression methods~\cite{chen2025hac} that focus solely on scene-specific priors, our approach generates a prior feature that integrates both scene-agnostic and scene-specific information.
Moreover, we employ an autoregressive network to generate context features from previously compressed elements within the attribute being encoded. Using an adaptive context selection algorithm with flexible receptive fields, we apply a larger receptive field in sparse areas to capture a broader context, and limit the context to the nearest $n$ elements in denser regions to focus on adjacent elements. This design enables more flexible autoregressive modeling for context feature generation than previous works~\cite{wang2024contextgs}.
Overall, by combining both the context and hyperprior features, we can generate a more accurate estimated distribution for each element, thereby enhancing entropy coding for improved lossless compression.

Overall, we integrate our HEMGS with several other coding components into an end-to-end optimized 3DGS data compression framework.
The experiment results on both Synthetic-NeRF~\cite{mildenhall2021nerf}, Tank\&Temples~\cite{knapitsch2017tanks}, Mip-NeRF360~\cite{barron2022mip}, and DeepBlending~\cite{hedman2018deep} datasets demonstrate that our newly-proposed compression method achieves the start-of-the-art compression performance, which achieves about 40\% average reduction in size over our baseline method while maintaining high rendering quality.
Our contributions can be summarized below:
    \textbf{1)} A Variable-rate Predictor that accepts a hyperparameter $\lambda$ to adjust storage and supports variable-rate lossy compression using a single model.
    \textbf{2)} A joint hyperprior and autoregressive model that produces a prior incorporating both scene-agnostic and scene-specific information, along with flexible context features for more accurate distribution estimation in entropy coding, thereby enhancing lossless compression.
    \textbf{3)} A hybrid entropy model, HEMGS, that integrates the above two components to achieve hybrid lossy-lossless compression of 3DGS data, saving about 40\% average storage while maintaining rendering performance compared to our baseline.



%% file: sec/2_relatedwork.tex
\section{Related Work}



\subsection{3D Gaussian Splatting Data Compression}

3D Gaussian Splatting (3DGS) represents 3D scenes using learnable shape and appearance attributes encoded as 3D Gaussian distributions. This method provides high-fidelity scene representation while enabling fast training and rendering, which has driven its widespread adoption. However, the large number of Gaussian points and their attributes present significant storage challenges, necessitating the development of effective 3DGS compression techniques.

Early compression methods focused on refining Gaussian parameter values to reduce model complexity. Approaches such as vector quantization grouped parameters into a pre-defined codebook~\cite{fan2023lightgaussian,lee2024compact,navaneet2023compact3d,niedermayr2024compressed}, while other techniques achieved compression by directly pruning parameters~\cite{fan2023lightgaussian,lee2024compact}.
Recent research~\cite{lu2024scaffold,morgenstern2023compact,chen2025hac,wang2024contextgs} has investigated structural relationships to improve compression efficiency. For instance, Scaffold-GS~\cite{lu2024scaffold} introduced anchor-centered features to reduce the number of 3DGS parameters. Building on Scaffold-GS, HAC~\cite{chen2025hac} employs a hash-grid to capture spatial relationships, while ContextGS~\cite{wang2024contextgs} integrates anchor-level contextual information to capture structural dependencies in 3DGS.


While these methods intuitively achieve 3DGS compression, the core of effective data compression lies in the entropy model.
A well-designed entropy model can uncover structural relationships in the data, enabling efficient coding and reducing storage requirements. However, current 3DGS compression methods still have room for improvement in entropy model design. Drawing inspiration from the success of entropy models in NIC methods \cite{balle2018variational,minnen2018joint,cheng2020learned} and NVC methods \cite{dvc,fvc,li2021dcvc}, we propose a hybrid entropy model for 3D Gaussian Splatting (HEMGS). This model integrates a hyperprior network to capture redundancies across attributes, an autoregressive network for redundancies within each attribute, and a variable-rate predictor to enable variable-rate compression in a single model. By combining these three modules, HEMGS effectively reduces redundancies, achieving high efficiency and greater versatility in 3DGS compression.

\subsection{Neural Entropy Model}

Entropy models are widely used in neural data compression to predict probability distributions efficiently for images, videos, or point clouds. For example, Minnen \textit{et al.} proposed an image compression method~\cite{minnen2018joint} that uses an entropy model combining a hyperprior network and an autoregressive network to capture spatial redundancies in images, achieving notable compression performance. Similar strategies have been applied to other neural data compression methods~\cite{cheng2020learned,5,6,9,dvc,fvc,li2021dcvc,han2024cra5,liu2023icmh,chen2022exploiting,chen2024group,chen2023neural,hu2020improving,Chen_2022_CVPR,liu2025efficient,10219641,liu2024towards}, highlighting the crucial role of entropy models in data compression.

Despite their effectiveness, entropy models in neural data compression methods are not directly applicable to 3DGS compression. A key challenge is that 3DGS contains multiple attributes, making it difficult for conventional entropy models to capture spatial redundancies across attributes. Additionally, the sparsity of anchor points in 3DGS limits the effectiveness of autoregressive networks used in NIC and NVC, as their receptive fields are insufficient for capturing meaningful information from sparse data.
To address these challenges, we propose a progressive coding algorithm within the hyperprior network. This method uses location features and previously compressed elements as priors to predict uncompressed attributes, effectively reducing redundancies across attributes. 
Additionally, our autoregressive model employs an adaptive context coding algorithm that adjusts the receptive field size according to the density of anchor points, enabling it to capture comprehensive and relevant context.
These innovations enhance the overall efficiency of 3DGS compression.

%% file: sec/3_methodology.tex
\section{Methodology}
\subsection{Preliminaries} \label{sec:pre}

\textbf{3D Gaussian Splatting (3DGS)} \cite{kerbl20233d} adopts a collection of Gaussians to represent 3D scenes, with each Gaussian initialized using Structure-from-Motion (SfM) and characterized by a 3D covariance matrix $\bm{\mathit{\Sigma}}$ and location $\boldsymbol{\mu}$:
\vspace{-3mm}
\begin{equation}
    G(\boldsymbol{x}) = e^{-\frac{1}{2}(\boldsymbol{x} - \boldsymbol{\mu})^\mathrm{T} \bm{\mathit{\Sigma}}^{-1} (\boldsymbol{x} - \boldsymbol{\mu})},
    \vspace{-3mm}
\end{equation}
where $\boldsymbol{x}$ denotes the coordinates of a 3D point. The covariance matrix $\bm{\mathit{\Sigma}}$ is expressed as $\bm{\mathit{\Sigma}} = \boldsymbol{R} \boldsymbol{S} \boldsymbol{S}^{\mathrm{T}} \boldsymbol{R}^{\mathrm{T}}$, where $\boldsymbol{R}$ and $\boldsymbol{S}$ represent the rotation and scaling matrices, respectively. Each Gaussian is also associated with opacity $\alpha$ and view-dependent color $\boldsymbol{c}$, which is modeled using Spherical Harmonics\cite{zhang2022differentiable}. To render an image, the 3D Gaussians are splatted to 2D, and the pixel values are computed using $\alpha$ and $\boldsymbol{c}$.

\textbf{Anchor} structure is introduced by Scaffold-GS~\cite{lu2024scaffold} to optimizes storage requirements. 
It employs anchors to group Gaussians and infers their attributes from the attributes of these anchors using MLPs. Each anchor is characterized by a location $\bm{x}^{a} \in \mathbb{R}^{3}$ and a set of attributes $\bm{\mathcal{A}}=\{\bm{f}^{a} \in \mathbb{R}^{D^{a}}, \bm{l} \in \mathbb{R}^{6}, \bm{o} \in \mathbb{R}^{3K}\}$, where $\bm{f}^{a}$ represents the anchor’s local feature, $\bm{l}$ represents the scaling factor, and $\bm{o}$ represents the offsets. During the rendering process, the anchor feature $\bm{f}^{a}$ is fed into the MLPs to produce the attributes for the Gaussians, whose positions are computed by combining $\bm{x}^{a}$ with $\bm{o}$, and $\bm{l}$ is used to regulate the positioning and shape of the Gaussians.

\textbf{Entropy Coding} requires the distribution of data $\bm{\mathcal{\hat{A}}}$ to encode it. However, the true distribution $q(\bm{\mathcal{\hat{A}}})$ of $\bm{\mathcal{\hat{A}}}$ is unknown and is typically approximated by an estimated distribution $p(\bm{\mathcal{\hat{A}}})$~\cite{balle2018variational,minnen2018joint,cheng2020learned,dvc,fvc,li2021dcvc}. According to \textit{Information Theory}~\cite{cover1999elements}, when entropy coding $\bm{\mathcal{\hat{A}}}$,
the cross-entropy $H(q,p)=\mathbb{E}_{\bm{\mathcal{\hat{A}}}\sim q}[-log(p(\bm{\mathcal{\hat{A}}}))]$ serves as the practical lower bound on the storage required. 
By refining $p(\bm{\mathcal{\hat{A}}})$ to better approximate $q(\bm{\mathcal{\hat{A}}})$, we can reduce $H(q,p)$ and thereby lower the storage cost.
This process models the conditional probability $p(\bm{\mathcal{\hat{A}}} | \cdot )$ in our compression framework based on the previously compressed content, and we use an arithmetic coding algorithm to perform the entropy coding procedure.

\begin{figure*}[htbp]
    \centering
    \includegraphics[width=\textwidth]{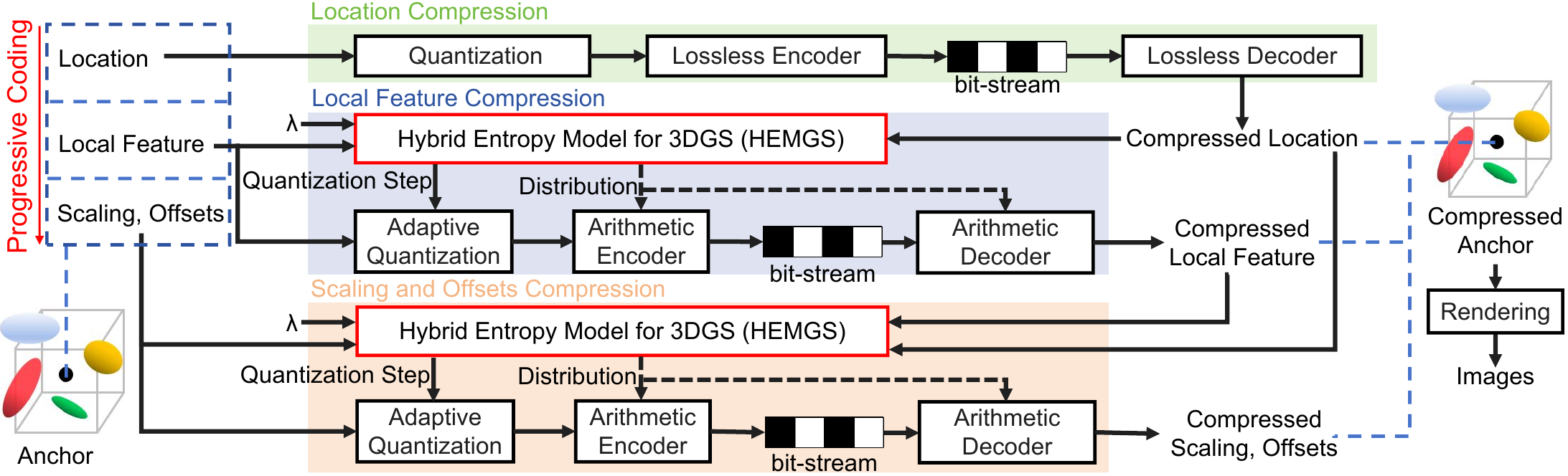}
    \vspace{-5mm}
    \caption{The overview of our 3DGS data compression framework, which integrates HEMGS with other additional coding components.}
    \label{fig:main_pipeline}
    \vspace{-4mm}
\end{figure*}

\subsection{Framework Overview} \label{sec:overview}
In anchor-based 3DGS methods~\cite{lu2024scaffold}, an anchor serves as an alternative representation by encapsulating both location information and various attributes, such as local features, scaling factors, and offsets. Therefore, compressing anchor data requires encoding both its location and these associated attributes.
To address this, we propose a compression framework based on our HEMGS that achieves hybrid lossy-lossless compression of those attributes, as illustrated in Figure~\ref{fig:main_pipeline}. Specifically, for each attribute, HEMGS applies adaptive quantization for lossy compression and estimates a Gaussian distribution for lossless entropy coding.
%
Overall, to exploit the dependency between anchor locations and their associated attributes, we adopt a \textit{progressive coding} procedure that sequentially compresses the anchor locations followed by the attributes. The details are presented below.

\textbf{Location Compression}
We begin by performing lossy compression on the anchor locations $\bm{x}^a$ through 16-bit quantization. The quantized location $\bm{\hat{x}}^a$ is then directly losslessly encoded into a bit-stream, which can later be losslessly decoded to reconstruct the compressed locations $\bm{\bar{x}}^a$. This process is a standard practice in many 3DGS data compression methods~\cite{chen2025hac,wang2024contextgs,lee2024compact,papantonakis2024reducing}.

\textbf{Local Feature Compression.}
Then, we compress the local feature $\bm{f}^a$ using HEMGS. First, HEMGS takes the compressed location feature $\bm{\bar{x}}^a$ and a hyperparameter $\lambda$ as inputs to generate a learned quantization step $\bm{s}_{\bm{f}}$ via the Variable-rate Predictor. This quantization step is then applied to the lossy compression of $\bm{f}^a$ using adaptive quantization, as described in HAC~\cite{chen2025hac}, yielding the quantized local feature $\bm{\hat{f}}^a$.
The adaptive quantization process is defined as
$\bm{\hat{f}}^a = \mathrm{Round}(\bm{f}^a/\bm{s}_{\bm{f}}) \times \bm{s}_{\bm{f}}$ and the with the storage adjusted according to the user-specified $\lambda$.

%
Then, for lossless compression of the quantized local feature $\bm{\hat{f}}^a$, our \textit{HEMGS} uses the previously compressed location $\bm{\bar{x}}^a$ as a prior and applies a joint autoregressive and hyperprior model to estimate the Gaussian distribution $p(\bm{\hat{f}}^a|\bm{\bar{x}}^a)$ for $\bm{\hat{f}}^a$. This distribution is used to perform lossless entropy coding with an \textit{arithmetic encoder}, which produces a bit-stream for storage or transmission. The \textit{arithmetic decoder} subsequently decodes the bit-stream using the predicted distribution $p(\bm{\hat{f}}^a|\bm{\bar{x}}^a)$, reconstructing the compressed attribute $\bm{\bar{f}}^a$.
More details of HEMGS can be found in Section~\ref{sec:HEMGS}.

\textbf{Scaling and Offsets Compression.} Next, we compress the scaling $\bm{l}$ and offsets $\bm{o}$ together by concatenating these two attributes into a single representation $(\bm{l}, \bm{o})$. Then, similar to the lossy-lossless compression procedure for the local feature, we input the compressed location $\bm{\bar{x}}^a$, the compressed local features $\bm{\bar{f}}^a$, and a user-specific $\lambda$ into HEMGS. HEMGS then produces a quantization step for performing adaptive quantization and lossy compression to produce $(\bm{l}, \bm{o})$, while subsequently estimating a distribution $p((\bm{\hat{l}}, \bm{\hat{o}})|\bm{\bar{x}}^a,\bm{\bar{f}}^a)$ for lossless compression (\textit{resp.,} decompression) via arithmetic encoding (\textit{resp.,} decoding) algorithm.

\textbf{Rendering.}
Last, the compressed location, local feature, scaling, and offsets together reassemble the compressed anchor. We will use all compressed anchors to render the novel-view image using the standard operations ~\cite{lu2024scaffold,chen2025hac,wang2024contextgs}.

\subsection{HEMGS for 3DGS Data Compression}
\label{sec:HEMGS}
The objective of our HEMGS is to enable hybrid lossy-lossless compression of anchors. It supports variable-rate lossy compression using a single model while incorporating enhanced neural entropy coding for lossless compression.
As illustrated in Figure~\ref{fig:main_pipeline}, when compressing the local feature $\bm{f}^a$, HEMGS takes as inputs the compressed location $\bm{\bar{x}}^a$ as prior information and a user-specified $\lambda$ to adjust storage. When compressing scaling $\bm{l}$ and offsets $\bm{o}$, the previously compressed local feature $\bm{\bar{f}}^a$ is also incorporated as an additional input alongside the aforementioned data. Specifically, HEMGS comprises three main components, a variable-rate predictor, a hyperprior network, and an autoregressive network, as detailed in Section\ref{sec:vrp}.

\begin{figure*}[htbp]
    \centering
    \includegraphics[width=\linewidth]{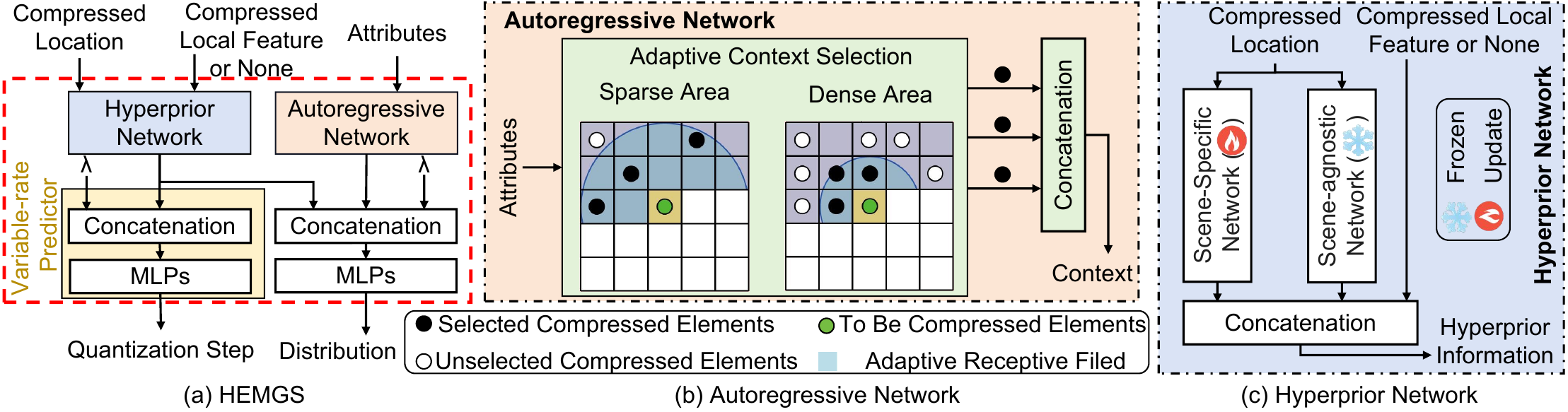}
    \vspace{-5mm}
    \caption{(a) The details of our HEMGS, which comprises a variable-rate predictor, a hyperprior network, and an autoregressive network. (b) The autoregressive network utilizes adaptive receptive fields to reduce redundancies within each attribute. (c) The hyperprior network incorporates scene-agnostic and scene-specific architectures to reduce redundancies across attributes.}
    \label{fig:network_details}
    \vspace{-5mm}
\end{figure*}

\subsubsection{Variable-rate Predictor} \label{sec:vrp}

Given the user-specific rate hyperparameter $\lambda$ and a prior feature derived from the compressed location $\bm{\bar{x}}^a$ (and the location feature $\bm{\bar{f}}^a$ when compressing the scaling and offset), we first expand $\lambda$ by duplicating its values to match the dimensionality of the prior feature. Next, we concatenate the expanded $\lambda$ with the prior feature and use MLPs to generate the learned quantization step feature $\bm{s}$.


This approach enables users to adjust storage requirements using a single model controlled by the variable hyperparameter  $\lambda$. Specifically, a larger $\lambda$ induces greater quantization, leading to reduced compressed attribute storage, whereas a smaller $\lambda$ results in less quantization and increased storage. In our implementation, we obtain an enhanced prior feature by directly leveraging the output of the hyperprior network, where provide the details in Section~\ref{sec:hyperprior}. 
%

\subsubsection{Hyperprior Network} \label{sec:hyperprior}

We introduce a hyperprior network to extract the previously compressed location and (or) attribute features, enabling more accurate distribution estimation for lossless entropy coding and reducing storage costs. 

\textbf{Scene-Agnostic and Scene-Specific Prior.} 
Different from existing methods that primarily capture scene-specific information in the prior feature through overfitting and storing a set of MLPs~\cite{lu2024scaffold,chen2025hac,wang2024contextgs}, our approach aims to generate a prior that incorporates both scene-agnostic and scene-specific network information for improved entropy coding.
Specifically, as illustrated in Figure~\ref{fig:network_details} (c), our scene-agnostic network leverages the capabilities of a pre-trained 3D feature extractor, such as PointNet++~\cite{qi2017pointnet++}, to effectively capture scene-agnostic structural relationships while introducing no additional storage overhead. Meanwhile, we directly employ an scene-specific network (\textit{e.g.}, hash-grid~\cite{chen2025hac})
to encode the structural relationships of individual scene-specific instances. By combining both scene-agnostic and scene-specific structural relationships, we produce a comprehensive hyperprior, which enables more accurate distribution estimation for anchor attributes, enhancing the efficiency of conditional entropy coding for local features. 

In combination with our progressive coding scheme, during the lossless compression of scaling and offsets, we further integrate the compressed local feature and leverage the features produced by both the scene-agnostic and scene-specific networks. This dual-path network generates a refined hyperprior feature that enables more effective entropy coding of the anchor attributes, namely local feature, scaling, and offsets, thereby enhancing lossless compression.

\subsubsection{Autoregressive Network}\label{sec:autoregressive}

Inspired by classic autoregressive coding mechanisms in NIC and NVC methods~\cite{minnen2018joint, cheng2020learned,5,6,9,li2021dcvc}, we have designed an autoregressive network specifically for 3DGS data to effectively generate context features from previously compressed elements within the attribute being encoded. This network employs a newly proposed adaptive context selection algorithm with a flexible receptive field to generate more comprehensive context feature for further distribution estimation, as illustrated in Figure~\ref{fig:network_details} (b).

Following existing anchor-based methods~\cite{lu2024scaffold,chen2025hac,wang2024contextgs}, which partition a 3D scene into voxels and designate the center of each non-empty voxel as an anchor point, we compress the anchors within these voxels in raster order. As each attribute is compressed, its elements can leverage surrounding, already-encoded elements as contextual information. Unlike previous 3DGS approaches that rely on fixed contextual cues~\cite{wang2024contextgs}, our design supports flexible, adaptive context selection for autoregressive modeling.

\textbf{Adaptive Context Selection.} 
We begin by defining a maximum receptive field size. If the number of elements within this field is below a threshold $n$,the encoded element is considered to be in a sparse area, and we retain all elements in the receptive field as contextual information to ensure sufficient context for accurate encoding. Conversely, if the number of elements exceeds $n$, the element is classified as residing in a dense area.
For dense regions, we sample the nearest $n$ elements to serve as contexts for the current encoding position. This restricts the receptive field to the minimal bounding region that contains these $n$ elements, thereby prioritizing the most relevant contextual information and ensuring that the most significant features are emphasized in high-density areas.

\subsection{Optimization}
The entire 3DGS data compression network can be optimized as below,
\vspace{-3mm}
\begin{equation}
    \mathcal{L} = \frac{1}{K}\sum_{k}^{K}(\mathcal{L}_{\mathrm{Rendering}} + \lambda_k\mathcal{L}_{\mathrm{anchor}}).
\vspace{-3mm}
\end{equation}
The term $\mathcal{L}_{\mathrm{Rendering}}$ is the rendering loss as specified in Scaffold-GS~\cite{lu2024scaffold}. The term $\mathcal{L}_{\mathrm{anchor}}$ denotes the estimated storage consumption for anchor, which is primarily consumed by the compressed attributes as mentioned previously, along with some auxiliary items used in the compression framework as in HAC~\cite{chen2025hac}. The details of $\mathcal{L}_{\mathrm{anchor}}$ are provided in the supplementary materials. To enable variable-rate encoding within a single model, $K$ different values of $\lambda$ are predefined as inputs to HEMGS, while each $\lambda$ also serves to balance the various loss components.




%% file: sec/4_experiment.tex
\begin{table*}[h]
\centering
\footnotesize
\setlength{\tabcolsep}{3pt}
\renewcommand{\arraystretch}{1.3}
\begin{tabular}{l|cccc|cccc|cccc|cccc}
\hline
\rowcolor[HTML]{FFFFFF} 
\textbf{Datasets}  & \multicolumn{4}{c|}{\textbf{Mip-NeRF360~\cite{barron2022mip}}} & \multicolumn{4}{c|}{\textbf{Tank\&Temples~\cite{knapitsch2017tanks}}} & \multicolumn{4}{c|}{\textbf{DeepBlending~\cite{hedman2018deep}}} & \multicolumn{4}{c}{\textbf{Synthetic-NeRF~\cite{mildenhall2021nerf}}} \\ \hline
\rowcolor[HTML]{FFFFFF} 
\textbf{Methods} & \textbf{psnr$\uparrow$} & \textbf{ssim$\uparrow$} & \textbf{lpips$\downarrow$} & \textbf{size$\downarrow$}  & \textbf{psnr$\uparrow$} & \textbf{ssim$\uparrow$} & \textbf{lpips$\downarrow$} & \textbf{size$\downarrow$}  & \textbf{psnr$\uparrow$} & \textbf{ssim$\uparrow$} & \textbf{lpips$\downarrow$} & \textbf{size$\downarrow$}  & \textbf{psnr$\uparrow$} & \textbf{ssim$\uparrow$} & \textbf{lpips$\downarrow$} & \textbf{size$\downarrow$} \\ \hline
3DGS~\cite{kerbl20233d} & 27.49 & \cellcolor{yellow!50}{0.813} & \cellcolor{pink}{0.222} & 744.7 & 23.69 & 0.844 & 0.178 & 431.0 & 29.42 & 0.899 & \cellcolor{pink}{0.247} & 663.9 & \cellcolor{yellow!50}{33.80} & \cellcolor{pink}{0.970} & \cellcolor{pink}{0.031} & 68.46 \\ 
Scaffold-GS~\cite{lu2024scaffold} & 27.50 & 0.806 & 0.252 & 253.9 & 23.96 & 0.853 & \cellcolor{yellow!50}{0.177} & 86.50 & 30.21 & 0.906 & 0.254 & 66.00 & 33.41 & 0.966 & 0.035 & 19.36 \\ \hline
EAGLES~\cite{girish2023eagles} & 27.15 & 0.808 & 0.238 & 68.89 & 23.41 & 0.840 & 0.200 & 34.00 & 29.91 & \cellcolor{yellow!50}{0.910} & \cellcolor{yellow!50}{0.250} & 62.00 & 32.54 & 0.965 & 0.039 & 5.74 \\ 
LightGaussian~\cite{fan2023lightgaussian} & 27.00 & 0.799 & 0.249 & 44.54 & 22.83 & 0.822 & 0.242 & 22.43 & 27.01 & 0.872 & 0.308 & 33.94 & 32.73 & 0.965 & 0.037 & 7.84 \\ 
Compact3DGS~\cite{lee2024compact} & 27.08 & 0.798 & 0.247 & 48.80 & 23.32 & 0.831 & 0.201 & 39.43 & 29.79 & 0.901 & 0.258 & 43.21 & 33.33 & \cellcolor{yellow!50}{0.968} & 0.034 & 5.54 \\ 
Compressed3D~\cite{navaneet2023compact3d} & 26.98 & 0.801 & 0.238 & 28.80 & 23.32 & 0.832 & 0.194 & 17.28 & 29.38 & 0.898 & 0.253 & 25.30 & 32.94& 0.967 & \cellcolor{yellow!50}{0.033} & 3.68 \\ 
Morgen. \textit{et al.}~\cite{morgenstern2023compact} & 26.01 & 0.772 & 0.259 & 23.90 & 22.78 & 0.817 & 0.211 & 13.05 & 28.92 & 0.891 & 0.276 & 8.40 & 31.05 & 0.955 & 0.047 & 2.20 \\ 
Navaneet \textit{et al.}~\cite{navaneet2023compact3d} & 27.16 & 0.808 & 0.228 & 50.30 & 23.47 & 0.840 & 0.188 & 27.97 & 29.75 & 0.903 & \cellcolor{pink}{0.247} & 42.77 & 33.09 & 0.967 & 0.036 & 4.42 \\ 
CompGS~\cite{liu2024compgs} & 27.26 & 0.803 & 0.239 & \cellcolor{yellow!50}{16.50} & 23.70 & 0.837 & 0.208 & 9.60 & 29.69 & 0.901 & 0.279 & 8.77 & - & - & - & - \\
HAC~\cite{chen2025hac}  &  \cellcolor{yellow!50}{27.77} & 0.811 & 0.230 & 21.87 & 24.40 & 0.853 & \cellcolor{yellow!50}{0.177} & 11.24 & 30.34 & 0.906 & 0.258 &  6.35 & 33.71 & \cellcolor{yellow!50}{0.968} & 0.034 & 1.86 \\ 
Context-GS~\cite{wang2024contextgs} & 27.72 & 0.811 & 0.231 & 21.58 & 24.29 & \cellcolor{yellow!50}{0.855} & \cellcolor{pink}{0.176} & 11.80 & \cellcolor{yellow!50}{30.39} & 0.909 & 0.258 & 6.60 & - & - & - & - \\ \hline
Ours (low-rate) & 27.68 & 0.809 & 0.239 & \cellcolor{pink}{12.52} & \cellcolor{yellow!50}{24.41} & 0.854 & 0.183 & \cellcolor{pink}{6.13} & 30.24 & 0.909 & 0.258 & \cellcolor{pink}{3.67} & 33.33 & 0.967 & 0.038 & \cellcolor{pink}{1.18} \\ 

Ours (high-rate) & \cellcolor{pink}{27.89} & \cellcolor{pink}{0.815} & \cellcolor{yellow!50}{0.226} &  17.56 & \cellcolor{pink}{24.62} & \cellcolor{pink}{0.857} & 0.180 & \cellcolor{yellow!50}{7.46} & \cellcolor{pink}{30.40} & \cellcolor{pink}{0.911} & 0.255 & \cellcolor{yellow!50}{4.25} & \cellcolor{pink}{33.82} & \cellcolor{yellow!50}{0.968} & 0.034 &  \cellcolor{yellow!50}{1.62}  \\ \hline
\end{tabular}
\vspace{-1mm}
\caption{Comparison of the newly proposed HEMGS method with other 3DGS data compression methods, including 3DGS and Scaffold-GS, for reference. For our approach, we report two sets of results by considering different trade-offs between size and fidelity, which are achieved by adjusting the parameter $\lambda$. The best and second-best results are highlighted in \colorbox{pink}{\textcolor{black}{red}} and \colorbox{yellow!50}{\textcolor{black}{yellow}} cells, respectively. The size values are measured in megabytes (MB).}
\label{tab:main_results}
\vspace{-2mm}
\end{table*}

\section{Experiment}
\subsection{Datasets}

We conducted comprehensive experiments on multiple datasets, including the small-scale Synthetic-NeRF dataset~\cite{mildenhall2021nerf} and three large-scale real-scene datasets: DeepBlending~\cite{hedman2018deep}, Mip-NeRF360~\cite{barron2022mip}, and Tanks\&Temples~\cite{knapitsch2017tanks}. 
The main paper reports the average results across all scenes for each dataset, while detailed quantitative results for individual scenes are presented in the supplementary material.

\subsection{Experiment Details}
\textbf{Benchmarks.} 
We compare our HEMGS with existing 3DGS data compression methods. Some methods~\cite{fan2023lightgaussian,lee2024compact,navaneet2023compact3d,niedermayr2024compressed} primarily achieve compact representations of 3DGS through parameter pruning or codebook-based strategies. Other methods~\cite{lu2024scaffold,morgenstern2023compact,chen2025hac,wang2024contextgs} exploit the structural relationships within 3DGS for compression. Notably, HAC~\cite{chen2025hac} and Context-GS~\cite{wang2024contextgs} have achieved impressive results in 3DGS compression by utilizing scene-specific hash-grids and anchor-level context to capture structural redundancies, respectively.

\textbf{Evaluation Metric.}
We measure the storage used for compressed 3DGS data using the metric of storage size in megabytes (MB). The quality of the rendered images from the compressed 3DGS data is evaluated using the Peak Signal-to-Noise Ratio (PSNR), Structural Similarity Index (SSIM)~\cite{wang2004ssim}, and Learned Perceptual Image Patch Similarity (LPIPS)~\cite{zhang2018LPIPS} metrics.

\textbf{Implementation Details.}
Our HEMGS is implemented in PyTorch with CUDA support. All experiments are conducted on an NVIDIA RTX3090 GPU (24GB memory). We use the Adam optimizer~\cite{kingma2014adam} and train our HEMGS for 60,000 iterations. 
To achieve variable-rate compression, we predefine four $\lambda$ values, including 1e-3, 2e-3, 3e-3, 4e-3, to optimize the entire compression framework within a single model.
In our autoregressive network, we set a large receptive field of $25\times25\times25$ at the voxel level and select the 20 closest elements within this field. The MLPs in our HEMGS consist of a single 2-layer MLP with ReLU activation.

\begin{figure*}[htbp]

    \begin{center}
        {
         \label{fig:sub_fig3}\includegraphics[width=0.32\textwidth]{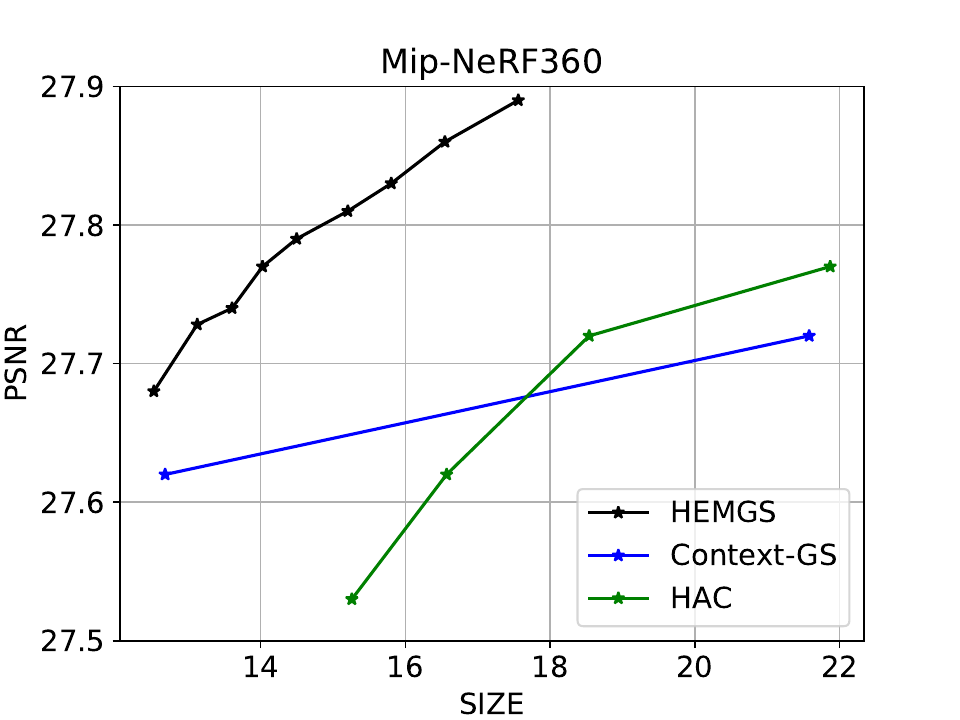}
        }
        {
         \label{fig:sub_fig2}\includegraphics[width=0.32\textwidth]{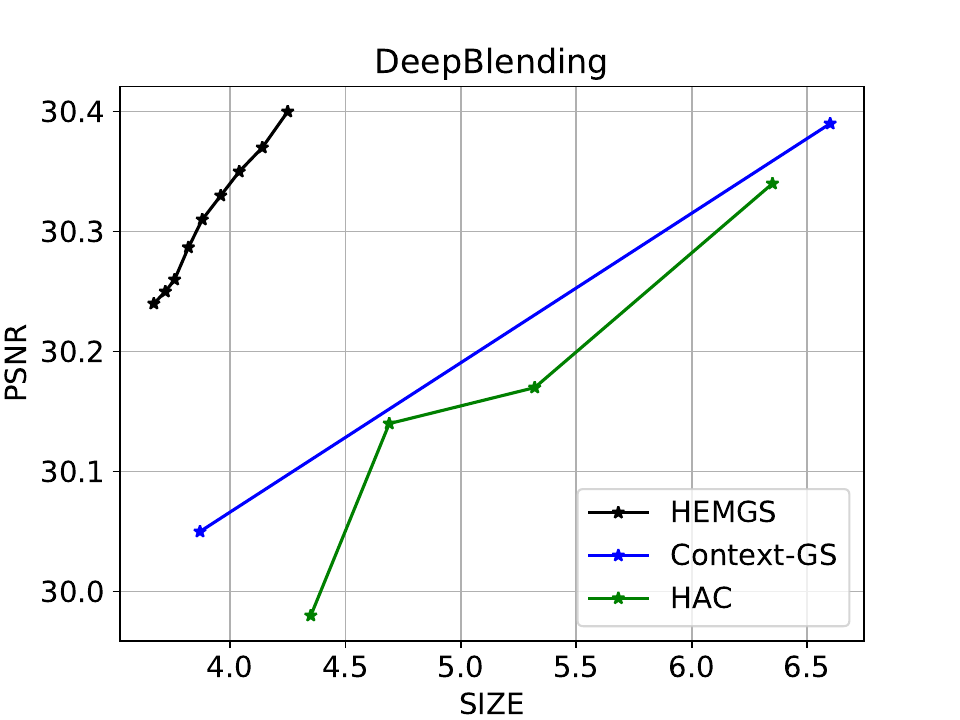}
        }
        {
         \label{fig:sub_fig1}\includegraphics[width=0.32\textwidth]{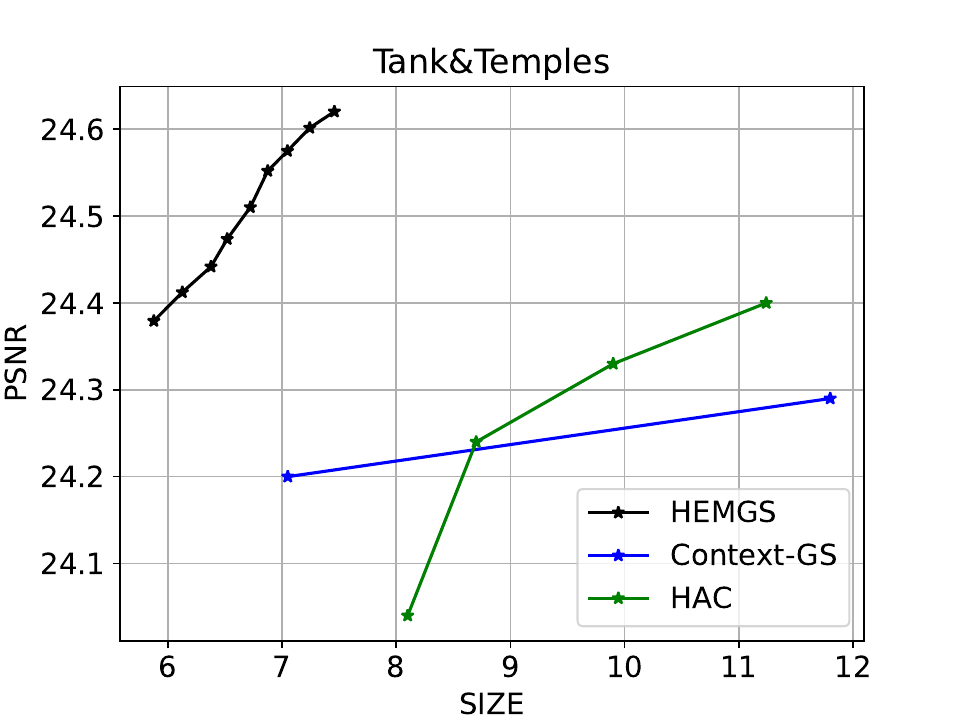}
        }
    \end{center}
    \vspace{-5mm}
    \caption{The Rate-Distortion (RD) curves on three benchmark datasets, including Mip-NeRF360, Tank\&Temples, and DeepBlending.}
    \label{fig:RD}
    \vspace{-3mm}
\end{figure*}

\subsection{Experiment Results}
The compression results are presented in Table~\ref{tab:main_results}. Our HEMGS achieves a substantial size reduction of over 100x compared to the 3DGS~\cite{kerbl20233d}, while maintaining improved fidelity. Additionally, it reduces size by over 18x on average when compared to Scaffold-GS~\cite{lu2024scaffold}. Furthermore, the proposed method demonstrates superior storage efficiency compared to recent state-of-the-art 3DGS data compression methods, such as HAC~\cite{chen2025hac} and Context-GS~\cite{wang2024contextgs}. The RD curves for different compression methods are provided in Figure~\ref{fig:RD}. 
Compared to our baseline method, HAC~\cite{chen2025hac}, our approach consistently achieves higher PSNR across all storage levels. Specifically, our HEMGS saves 46.43\% storage at 24.4 PSNR on the Tank\&Temples dataset~\cite{knapitsch2017tanks} and 39.34\% storage at 30.3 PSNR on the DeepBlending dataset~\cite{hedman2018deep} compared to HAC.
Overall, our approach achieves an average storage savings of approximately 40\% across all benchmark datasets, including Mip-NeRF360~\cite{barron2022mip}, Tank\&Temples~\cite{knapitsch2017tanks}, DeepBlending~\cite{hedman2018deep}, and Synthetic-NeRF~\cite{mildenhall2021nerf}. We also report the BDBR~\cite{bdbr} results in Table~\ref{tab:bdbr}, showing that HAC and Context-GS require 52.52\% and 66.47\% more storage, respectively, than our newly proposed HEMGS at the same PSNR on the Mip-NeRF360 dataset.
This demonstrates that our method outperforms the recent state-of-the-art approach, HAC, and highlights the efficiency of our newly proposed HEMGS for 3DGS data compression.


\begin{table}[ht]
    \centering
    \vspace{-2mm}
    \begin{tabular}{c|cc}
    \hline
        & Mip-NeRF360 & DeepBlending \\ \hline
         Context-GS~\cite{wang2024contextgs} & 66.47 & 42.03  \\ \hline
         HAC~\cite{chen2025hac} & 52.52 & 82.85  \\ \hline
    \end{tabular}
    \vspace{-2mm}
    \caption{BDBR (\%) results for HAC, Context-GS, and our newly proposed HEMGS across different datasets. Positive BDBR values indicate additional storage costs than our HEMGS.}
    \label{tab:bdbr}
    \vspace{-1mm}
\end{table}

\subsection{Ablation Study and Analysis}
\textbf{Ablation of Scene-agnostic Network and Autoregressive Network.} In this subsection, we conduct ablation studies to evaluate the effectiveness of scene-agnostic network in the hyperprior network and the autoregressive network. Experiments are performed on the ``playroom'' scene in the DeepBlending dataset~\cite{hedman2018deep}, as shown in Table~\ref{tab:ab}. To assess the scene-agnostic network, we remove it from ``Ours'' (denoted as ``Ours w/o SA''), leading to a 0.26\% PSNR decrease and an additional 6.95\% storage cost. For the autoregressive network, we further remove it from ``Ours w/o SA'' (denoted as ``Ours w/o SA, and AR''), resulting in a 0.16\% PSNR decrease and an additional 16.10\% storage cost. These results demonstrate the effectiveness of our Scene-Agnostic network, and autoregressive network in enhancing PSNR while reducing storage requirements.

\begin{table}[t]
    \centering
    \vspace{-2mm}
    \begin{tabular}{l|cc} \hline

      Methods & PSNR$\uparrow$ & Size (MB)$\downarrow$  \\ \hline
        Ours &  30.75 & 3.31  \\ 
         Ours w/o SA & 30.67 & 3.54 \\ 
         Ours w/o SA, and AR & 30.62 & 4.11  \\ \hline
    \end{tabular}
    \vspace{-2mm}
    \caption{Ablation study of each newly proposed component, which is measured on the ``playroom'' scene from the DeepBlending dataset.  ``SA'', and ``AR'' denote \textbf{S}cene-\textbf{A}gnostic network, and \textbf{A}uto\textbf{R}egressive network, respectively.}
    \label{tab:ab}
    \vspace{-6mm}
\end{table}


\textbf{Ablation of Adaptive Context Selection.} To measure the impact of our adaptive context selection algorithm and large receptive field in our autoregressive network, we conduct an ablation study, as shown in Table~\ref{tab:ab_2}. To evaluate the effectiveness of our adaptive context selection algorithm, we remove this component from our method, namely ``Ours w/o ACS''. The results for ``Ours w/o ACS'' show an additional 0.19 MB storage cost compared to our method. To assess the impact of a large receptive field ($25\times25\times25$), we reduce it to $5\times5\times5$ as in NIC methods, resulting in ``Ours with SRF''. Notably, the small receptive field captures an average of only 0.43 anchor points, significantly fewer than the 5.76 anchor points obtained by our method.
This change in ``Ours with SRF'' incurs an additional 0.26 MB storage cost and a 0.02 dB PSNR drop compared to ``Ours w/o ACS''. These results demonstrate the effectiveness of our adaptive context selection algorithm and large receptive field.

\begin{table}[t]
    \centering
    \vspace{-2mm}
    \begin{tabular}{l|cc|cc} \hline
        \multirow{2}{*}{Methods}& \multicolumn{2}{c|}{Anchor Number} & \multicolumn{2}{c}{Performance} \\ \cline{2-5}
       & Average & Max & PSNR$\uparrow$ & Size $\downarrow$  \\ \hline
         Ours & 5.76 & 20 & 30.67 & 3.54 \\ 
         Ours w/o ACS & 6.15 & 80 & 30.67 & 3.75 \\ 
         Ours with SRF & 0.43 & 30 & 30.65 & 4.01  \\ \hline
    \end{tabular}
    \vspace{-2mm}
    \caption{Ablation study of the adaptive context selection algorithm and large receptive field, two components of the autoregressive network, on the ``playroom'' scene from the DeepBlending dataset. The average and maximum anchor numbers in the receptive field along with the performance of the compressed 3D scene are reported. ``ACS'', and ``SRF'' represent the \textbf{A}daptive \textbf{C}ontext \textbf{S}election algorithm, and \textbf{S}mall \textbf{R}eceptive \textbf{F}iled ($5\times5\times5$) as in the NIC methods, respectively.}
    \label{tab:ab_2}
    \vspace{-3mm}
\end{table}

\begin{table*}
    \centering
    \begin{tabular}{c|cccccc|cc} \hline
    \multirow{2}{*}{Methods}& \multicolumn{6}{c|}{Storage Costs (MB)$\downarrow$} & \multicolumn{2}{c}{Fidlity$\uparrow$} \\ \cline{2-9}
         & Location & Feature & Scaling & Offsets & Others & Total & PSNR & SSIM \\ \hline
         HAC & 0.897 & 1.366 & 0.716 & 0.974 & 0.362 & 4.318 & 30.22 & 0.908 \\
         Ours & 0.736 & 1.244 & 0.648 & 0.640 & 0.402 & 3.670 & 30.24 & 0.909 \\ \hline
    \end{tabular}
    \vspace{-2mm}
    \caption{The storage cost of each component and the rendering qualities of HAC and our newly proposed HEMGS on the DeepBlending dataset. ``Others'' refers to additional storage costs (\textit{e.g.}, the parameters of MLPs), detailed in the supplementary materials.}
    \label{tab:storage_cost}
    \vspace{-4mm}
\end{table*}

\textbf{Analysis of the Storage Cost.} To further evaluate the effectiveness of our HEMGS, we compared the storage requirements for anchor locations and various attributes, as shown in Table~\ref{tab:storage_cost}. Our method reduces the storage needed for locations. For example, our approach achieves 18\% reduction in location storage compared to HAC~\cite{chen2025hac}, demonstrating more efficient location compression. Additionally, our method decreases storage for anchor local features, achieving a 9\% reduction compared to HAC. This improvement is due to the use of both scene-agnostic and scene-specific architectures to extract location features, combined with contextual information from the autoregressive model to predict local feature distributions, enabling efficient encoding. Furthermore, our method yields significant storage reductions for scaling and offset attributes. Storage is reduced by 9\% and 34\% for scaling and offset, respectively, compared to HAC. The greater reduction in scaling and offset is primarily due to using compressed features as prior information to predict the distribution of these attributes, effectively reducing structural redundancies. This validates the effectiveness of our progressive coding algorithm in reducing attribute redundancy. Overall, compared to the baseline method HAC~\cite{chen2025hac}, our approach not only reduces storage consumption across most components but also achieves higher image fidelity, confirming the efficiency of our method.

\begin{figure}
    \centering
    \vspace{-2mm}
    \includegraphics[width=0.7\linewidth]{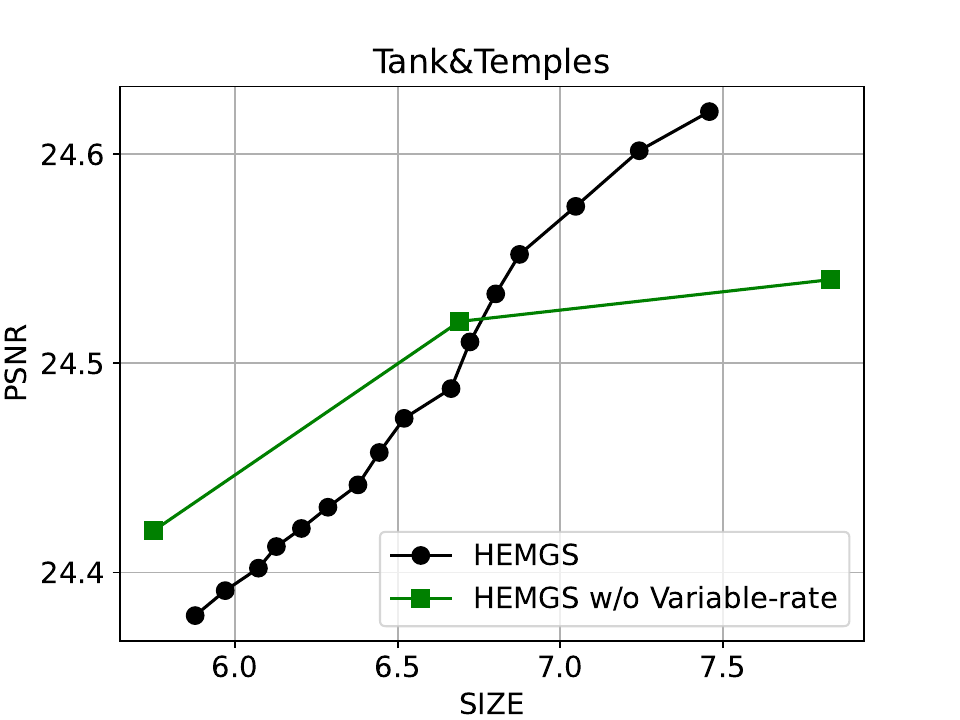}
    \vspace{-2mm}
    \caption{The Rate-Distortion (RD) curves of our variable-rate HEMGS method for compressing 3DGS data using a single model on the Tank\&Temples dataset. ``HEMGS w/o Variable-rate'' represents our HEMGS without using Variable-rate Predictor, which needs a separate model for each rate point.}
    \label{fig:variable_rate}
\end{figure}

\textbf{Ablation of Variable-rate Predictor.} The variable-rate results are presented in Figure~\ref{fig:variable_rate}. Across 16 rate points, our single HEMGS model demonstrates effective variable-rate adjustment while achieving performance comparable to HEMGS without using the variable-rate predictor. 
Moreover, it is worth noting that, unlike HEMGS, other compression methods (\textit{e.g.}, HAC~\cite{chen2025hac}, Context-GS~\cite{wang2024contextgs}) not only require training multiple models for different storage but also exhibit slightly inferior compression performance.


\begin{table}[t]
    \centering
    \setlength{\tabcolsep}{4pt}
    \vspace{-2mm}
    \begin{tabular}{l|c|c}
    \hline
         Methods & Training Time (Hours) & Model Size (MB) \\ \hline
         Context-GS & $5\times N$ & $0.316 \times N$ \\  
         HAC & $0.5\times N$ & $0.157 \times N$ \\  
         Ours & $1\times1$& $0.236\times 1$ \\ \hline
    \end{tabular}
    \vspace{-2mm}
    \caption{The complexity experiments on the DeepBlending dataset. $N$ represents the number of rate points, which is typically larger in real-world applications.}
    \vspace{-5mm}
    \label{tab:time}
\end{table}

\textbf{Analysis of the Complexity.} We conduct the complexity experiments in training time and model size with the previous 3DGS compression methods Context-GS~\cite{wang2024contextgs} and HAC~\cite{chen2025hac}, as shown in Table~\ref{tab:time}. For training time, we measure the time on the NVIDIA RTX3090 GPU. As aforementioned, our HEMGS enable variable-rate compression in a single model. 
Given that different scenarios require varying levels of compression, a versatile compression method should support multiple rates. Previous approaches, such as HAC~\cite{chen2025hac} and Context-GS~\cite{wang2024contextgs}, require retraining a separate model for each rate, significantly increasing training and storage costs.
As the demand for variable-rate compression increases, our method offers greater advantages by reducing both training time and model storage requirements.

\begin{figure}[t!]
    \centering
    \vspace{-2mm}
    \includegraphics[width=0.8\linewidth]{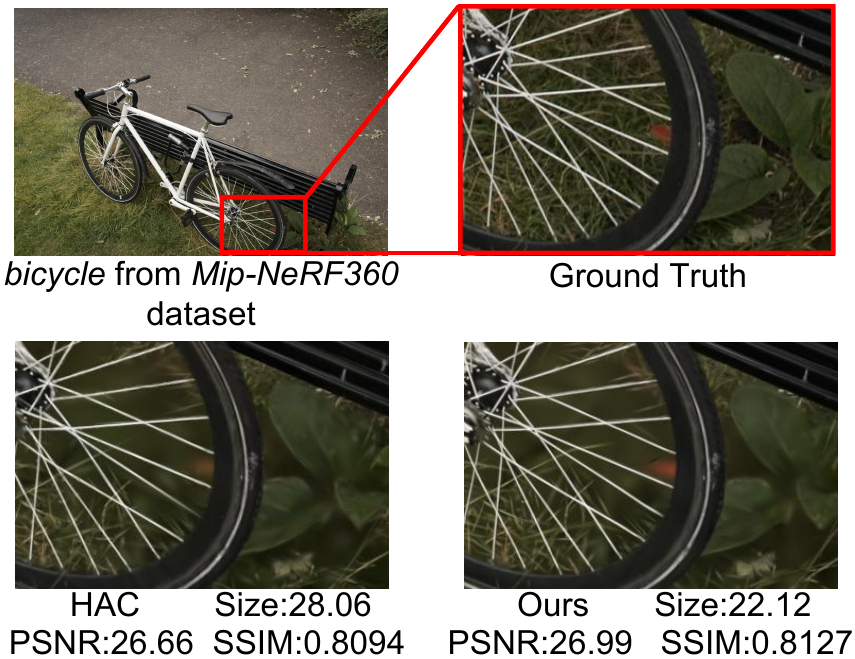}
    \vspace{-2mm}
    \caption{Visualization comparison between HAC and our newly proposed HEMGS on the ``bicycle'' scene from the Mip-NeRF360 dataset. PSNR, and SSIM of the rendered image as well as the size (MB) of the scene are reported.}
    \label{fig:visualization}
    \vspace{-5mm}
\end{figure}

\textbf{Visualization.}
Our visualization results are displayed in Figure~\ref{fig:visualization}, showing rendered outputs for the ``bicycle'' scene from the Mip-NeRF360 dataset~\cite{barron2022mip}. 
In the ``bicycle'' scene, images rendered by our method display more distinct details, especially in the central region of the image, compared to HAC. Metrics including file size, PSNR, and SSIM consistently show that our approach preserves greater image fidelity with reduced storage requirements. These visual comparisons effectively highlight the efficiency of our HEMGS approach.



\section{Conclusion}
In this work, we have proposed HEMGS, a novel 3DGS data compression approach that enables effective variable-rate compression using a single model. Our method offers a robust solution for real-world applications through a powerful joint entropy model, while achieving state-of-the-art performance across multiple benchmarks. Our work also paves the way for further research into comprehensive models that address a wide range of compression scenarios, from extreme to high-quality compression.

\textbf{Limitations and Future Works}. Currently, our method HEMGS applies only anchor-based 3DGS method, which requires the anchor structure to distribute local 3D Gaussians. In our future work, we plan to extend it to anchor-free data structures to broaden its applicability.






%% file: main.bbl
\begin{thebibliography}{42}
\providecommand{\natexlab}[1]{#1}
\providecommand{\url}[1]{\texttt{#1}}
\expandafter\ifx\csname urlstyle\endcsname\relax
  \providecommand{\doi}[1]{doi: #1}\else
  \providecommand{\doi}{doi: \begingroup \urlstyle{rm}\Url}\fi

\bibitem[Ball{\'e} et~al.(2018)Ball{\'e}, Minnen, Singh, Hwang, and Johnston]{balle2018variational}
Johannes Ball{\'e}, David Minnen, Saurabh Singh, Sung~Jin Hwang, and Nick Johnston.
\newblock Variational image compression with a scale hyperprior.
\newblock \emph{ICLR}, 2018.

\bibitem[Barron et~al.(2022)Barron, Mildenhall, Verbin, Srinivasan, and Hedman]{barron2022mip}
Jonathan~T Barron, Ben Mildenhall, Dor Verbin, Pratul~P Srinivasan, and Peter Hedman.
\newblock Mip-nerf 360: Unbounded anti-aliased neural radiance fields.
\newblock In \emph{CVPR}, pages 5470--5479, 2022.

\bibitem[Chen et~al.(2025)Chen, Wu, Lin, Harandi, and Cai]{chen2025hac}
Yihang Chen, Qianyi Wu, Weiyao Lin, Mehrtash Harandi, and Jianfei Cai.
\newblock Hac: Hash-grid assisted context for 3d gaussian splatting compression.
\newblock In \emph{ECCV}, pages 422--438. Springer, 2025.

\bibitem[Chen et~al.(2022{\natexlab{a}})Chen, Gu, Lu, and Xu]{chen2022exploiting}
Zhenghao Chen, Shuhang Gu, Guo Lu, and Dong Xu.
\newblock Exploiting intra-slice and inter-slice redundancy for learning-based lossless volumetric image compression.
\newblock \emph{IEEE TIP}, 31:\penalty0 1697--1707, 2022{\natexlab{a}}.

\bibitem[Chen et~al.(2022{\natexlab{b}})Chen, Lu, Hu, Liu, Jiang, and Xu]{Chen_2022_CVPR}
Zhenghao Chen, Guo Lu, Zhihao Hu, Shan Liu, Wei Jiang, and Dong Xu.
\newblock Lsvc: A learning-based stereo video compression framework.
\newblock In \emph{CVPR}, pages 6073--6082, 2022{\natexlab{b}}.

\bibitem[Chen et~al.(2023)Chen, Relic, Azevedo, Zhang, Gross, Xu, Zhou, and Schroers]{chen2023neural}
Zhenghao Chen, Lucas Relic, Roberto Azevedo, Yang Zhang, Markus Gross, Dong Xu, Luping Zhou, and Christopher Schroers.
\newblock Neural video compression with spatio-temporal cross-covariance transformers.
\newblock In \emph{ACM MM}, pages 8543--8551, 2023.

\bibitem[Chen et~al.(2024)Chen, Zhou, Hu, and Xu]{chen2024group}
Zhenghao Chen, Luping Zhou, Zhihao Hu, and Dong Xu.
\newblock Group-aware parameter-efficient updating for content-adaptive neural video compression.
\newblock In \emph{ACM MM}, pages 11022--11031, 2024.

\bibitem[Cheng et~al.(2020)Cheng, Sun, Takeuchi, and Katto]{cheng2020learned}
Zhengxue Cheng, Heming Sun, Masaru Takeuchi, and Jiro Katto.
\newblock Learned image compression with discretized gaussian mixture likelihoods and attention modules.
\newblock In \emph{Proceedings of the IEEE/CVF conference on computer vision and pattern recognition}, pages 7939--7948, 2020.

\bibitem[Cover(1999)]{cover1999elements}
Thomas~M Cover.
\newblock \emph{Elements of information theory}.
\newblock John Wiley \& Sons, 1999.

\bibitem[Fan et~al.(2023)Fan, Wang, Wen, Zhu, Xu, and Wang]{fan2023lightgaussian}
Zhiwen Fan, Kevin Wang, Kairun Wen, Zehao Zhu, Dejia Xu, and Zhangyang Wang.
\newblock Lightgaussian: Unbounded 3d gaussian compression with 15x reduction and 200+ fps.
\newblock \emph{arXiv preprint arXiv:2311.17245}, 2023.

\bibitem[Girish et~al.(2023)Girish, Gupta, and Shrivastava]{girish2023eagles}
Sharath Girish, Kamal Gupta, and Abhinav Shrivastava.
\newblock Eagles: Efficient accelerated 3d gaussians with lightweight encodings.
\newblock \emph{arXiv preprint arXiv:2312.04564}, 2023.

\bibitem[Gisle(2001)]{bdbr}
Bjontegaard Gisle.
\newblock Calculation of average psnr differences between rd curves.
\newblock In \emph{VCEG-M33}, page~7, 2001.

\bibitem[Han et~al.(2024)Han, Chen, Guo, Xu, and Bai]{han2024cra5}
Tao Han, Zhenghao Chen, Song Guo, Wanghan Xu, and Lei Bai.
\newblock Cra5: Extreme compression of era5 for portable global climate and weather research via an efficient variational transformer.
\newblock \emph{arXiv preprint arXiv:2405.03376}, 2024.

\bibitem[He et~al.(2021)He, Zheng, Sun, Wang, and Qin]{6}
Dailan He, Yaoyan Zheng, Baocheng Sun, Yan Wang, and Hongwei Qin.
\newblock Checkerboard context model for efficient learned image compression.
\newblock In \emph{CVPR}, pages 14771--14780, 2021.

\bibitem[Hedman et~al.(2018)Hedman, Philip, Price, Frahm, Drettakis, and Brostow]{hedman2018deep}
Peter Hedman, Julien Philip, True Price, Jan-Michael Frahm, George Drettakis, and Gabriel Brostow.
\newblock Deep blending for free-viewpoint image-based rendering.
\newblock \emph{ACM Trans. Graph.}, 37\penalty0 (6):\penalty0 1--15, 2018.

\bibitem[Hu et~al.(2020)Hu, Chen, Xu, Lu, Ouyang, and Gu]{hu2020improving}
Zhihao Hu, Zhenghao Chen, Dong Xu, Guo Lu, Wanli Ouyang, and Shuhang Gu.
\newblock Improving deep video compression by resolution-adaptive flow coding.
\newblock In \emph{ECCV}, pages 193--209. Springer, 2020.

\bibitem[Hu et~al.(2021)Hu, Lu, and Xu]{fvc}
Zhihao Hu, Guo Lu, and Dong Xu.
\newblock Fvc: A new framework towards deep video compression in feature space.
\newblock In \emph{CVPR}, pages 1502--1511, 2021.

\bibitem[Kerbl et~al.(2023)Kerbl, Kopanas, Leimk{\"u}hler, and Drettakis]{kerbl20233d}
Bernhard Kerbl, Georgios Kopanas, Thomas Leimk{\"u}hler, and George Drettakis.
\newblock 3d gaussian splatting for real-time radiance field rendering.
\newblock \emph{ACM Trans. Graph.}, 42\penalty0 (4):\penalty0 139--1, 2023.

\bibitem[Kingma(2015)]{kingma2014adam}
Diederik~P Kingma.
\newblock Adam: A method for stochastic optimization.
\newblock \emph{ICLR}, 2015.

\bibitem[Knapitsch et~al.(2017)Knapitsch, Park, Zhou, and Koltun]{knapitsch2017tanks}
Arno Knapitsch, Jaesik Park, Qian-Yi Zhou, and Vladlen Koltun.
\newblock Tanks and temples: Benchmarking large-scale scene reconstruction.
\newblock \emph{ACM Trans. Graph.}, 36\penalty0 (4):\penalty0 1--13, 2017.

\bibitem[Lee et~al.(2024)Lee, Rho, Sun, Ko, and Park]{lee2024compact}
Joo~Chan Lee, Daniel Rho, Xiangyu Sun, Jong~Hwan Ko, and Eunbyung Park.
\newblock Compact 3d gaussian representation for radiance field.
\newblock In \emph{CVPR}, pages 21719--21728, 2024.

\bibitem[Li et~al.(2021)Li, Li, and Lu]{li2021dcvc}
Jiahao Li, Bin Li, and Yan Lu.
\newblock Deep contextual video compression.
\newblock \emph{Advances in Neural Information Processing Systems}, 34, 2021.

\bibitem[Liu et~al.(2023{\natexlab{a}})Liu, Hu, Chen, and Xu]{liu2023icmh}
Lei Liu, Zhihao Hu, Zhenghao Chen, and Dong Xu.
\newblock Icmh-net: Neural image compression towards both machine vision and human vision.
\newblock In \emph{ACM MM}, pages 8047--8056, 2023{\natexlab{a}}.

\bibitem[Liu et~al.(2023{\natexlab{b}})Liu, Hu, and Zhang]{10219641}
Lei Liu, Zhihao Hu, and Jing Zhang.
\newblock Pchm-net: A new point cloud compression framework for both human vision and machine vision.
\newblock In \emph{ICME}, pages 1997--2002, 2023{\natexlab{b}}.

\bibitem[Liu et~al.(2024{\natexlab{a}})Liu, Hu, and Chen]{liu2024towards}
Lei Liu, Zhihao Hu, and Zhenghao Chen.
\newblock Towards point cloud compression for machine perception: A simple and strong baseline by learning the octree depth level predictor.
\newblock In \emph{International Joint Conference on Artificial Intelligence WorkShop}, pages 3--17. Springer, 2024{\natexlab{a}}.

\bibitem[Liu et~al.(2025)Liu, Chen, Hu, and Xu]{liu2025efficient}
Lei Liu, Zhenghao Chen, Zhihao Hu, and Dong Xu.
\newblock An efficient adaptive compression method for human perception and machine vision tasks.
\newblock \emph{arXiv preprint arXiv:2501.04329}, 2025.

\bibitem[Liu et~al.(2024{\natexlab{b}})Liu, Wu, Zhang, Wang, Li, and Kwong]{liu2024compgs}
Xiangrui Liu, Xinju Wu, Pingping Zhang, Shiqi Wang, Zhu Li, and Sam Kwong.
\newblock Compgs: Efficient 3d scene representation via compressed gaussian splatting.
\newblock In \emph{ACM MM}, 2024{\natexlab{b}}.

\bibitem[Lu et~al.(2019)Lu, Ouyang, Xu, Zhang, Cai, and Gao]{dvc}
Guo Lu, Wanli Ouyang, Dong Xu, Xiaoyun Zhang, Chunlei Cai, and Zhiyong Gao.
\newblock Dvc: An end-to-end deep video compression framework.
\newblock In \emph{CVPR}, pages 11006--11015, 2019.

\bibitem[Lu et~al.(2024)Lu, Yu, Xu, Xiangli, Wang, Lin, and Dai]{lu2024scaffold}
Tao Lu, Mulin Yu, Linning Xu, Yuanbo Xiangli, Limin Wang, Dahua Lin, and Bo Dai.
\newblock Scaffold-gs: Structured 3d gaussians for view-adaptive rendering.
\newblock In \emph{CVPR}, pages 20654--20664, 2024.

\bibitem[Mildenhall et~al.(2021)Mildenhall, Srinivasan, Tancik, Barron, Ramamoorthi, and Ng]{mildenhall2021nerf}
Ben Mildenhall, Pratul~P Srinivasan, Matthew Tancik, Jonathan~T Barron, Ravi Ramamoorthi, and Ren Ng.
\newblock Nerf: Representing scenes as neural radiance fields for view synthesis.
\newblock \emph{Communications of the ACM}, 65\penalty0 (1):\penalty0 99--106, 2021.

\bibitem[Minnen et~al.(2018)Minnen, Ball{\'e}, and Toderici]{minnen2018joint}
David Minnen, Johannes Ball{\'e}, and George~D Toderici.
\newblock Joint autoregressive and hierarchical priors for learned image compression.
\newblock \emph{NeurIPS}, 31, 2018.

\bibitem[Morgenstern et~al.(2025)Morgenstern, Barthel, Hilsmann, and Eisert]{morgenstern2023compact}
Wieland Morgenstern, Florian Barthel, Anna Hilsmann, and Peter Eisert.
\newblock Compact 3d scene representation via self-organizing gaussian grids.
\newblock \emph{ECCV}, 2025.

\bibitem[Navaneet et~al.(2023)Navaneet, Meibodi, Koohpayegani, and Pirsiavash]{navaneet2023compact3d}
KL Navaneet, Kossar~Pourahmadi Meibodi, Soroush~Abbasi Koohpayegani, and Hamed Pirsiavash.
\newblock Compact3d: Compressing gaussian splat radiance field models with vector quantization.
\newblock \emph{arXiv preprint arXiv:2311.18159}, 2023.

\bibitem[Niedermayr et~al.(2024)Niedermayr, Stumpfegger, and Westermann]{niedermayr2024compressed}
Simon Niedermayr, Josef Stumpfegger, and R{\"u}diger Westermann.
\newblock Compressed 3d gaussian splatting for accelerated novel view synthesis.
\newblock In \emph{CVPR}, pages 10349--10358, 2024.

\bibitem[Papantonakis et~al.(2024)Papantonakis, Kopanas, Kerbl, Lanvin, and Drettakis]{papantonakis2024reducing}
Panagiotis Papantonakis, Georgios Kopanas, Bernhard Kerbl, Alexandre Lanvin, and George Drettakis.
\newblock Reducing the memory footprint of 3d gaussian splatting.
\newblock \emph{Proceedings of the ACM on Computer Graphics and Interactive Techniques}, 7\penalty0 (1):\penalty0 1--17, 2024.

\bibitem[Qi et~al.(2017)Qi, Yi, Su, and Guibas]{qi2017pointnet++}
Charles~Ruizhongtai Qi, Li Yi, Hao Su, and Leonidas~J Guibas.
\newblock Pointnet++: Deep hierarchical feature learning on point sets in a metric space.
\newblock \emph{NeurIPS}, 30, 2017.

\bibitem[Qian et~al.(2021)Qian, Tan, Sun, Lin, Li, Sun, Hao, and Jin]{9}
Yichen Qian, Zhiyu Tan, Xiuyu Sun, Ming Lin, Dongyang Li, Zhenhong Sun, Li Hao, and Rong Jin.
\newblock Learning accurate entropy model with global reference for image compression.
\newblock In \emph{ICLR}, 2021.

\bibitem[Qian et~al.(2022)Qian, Sun, Lin, Tan, and Jin]{5}
Yichen Qian, Xiuyu Sun, Ming Lin, Zhiyu Tan, and Rong Jin.
\newblock Entroformer: A transformer-based entropy model for learned image compression.
\newblock In \emph{ICLR}, 2022.

\bibitem[Wang et~al.(2024)Wang, Li, Guo, Yang, Kot, and Wen]{wang2024contextgs}
Yufei Wang, Zhihao Li, Lanqing Guo, Wenhan Yang, Alex Kot, and Bihan Wen.
\newblock Context{GS} : Compact 3d gaussian splatting with anchor level context model.
\newblock In \emph{The Thirty-eighth Annual Conference on Neural Information Processing Systems}, 2024.

\bibitem[Wang et~al.(2004)Wang, Bovik, Sheikh, and Simoncelli]{wang2004ssim}
Zhou Wang, Alan~C Bovik, Hamid~R Sheikh, and Eero~P Simoncelli.
\newblock Image quality assessment: from error visibility to structural similarity.
\newblock \emph{IEEE TIP}, 13\penalty0 (4):\penalty0 600--612, 2004.

\bibitem[Zhang et~al.(2022)Zhang, Baek, Rusinkiewicz, and Heide]{zhang2022differentiable}
Qiang Zhang, Seung-Hwan Baek, Szymon Rusinkiewicz, and Felix Heide.
\newblock Differentiable point-based radiance fields for efficient view synthesis.
\newblock In \emph{SIGGRAPH Asia 2022 Conference Papers}, pages 1--12, 2022.

\bibitem[Zhang et~al.(2018)Zhang, Isola, Efros, Shechtman, and Wang]{zhang2018LPIPS}
Richard Zhang, Phillip Isola, Alexei~A Efros, Eli Shechtman, and Oliver Wang.
\newblock The unreasonable effectiveness of deep features as a perceptual metric.
\newblock In \emph{CVPR}, pages 586--595, 2018.

\end{thebibliography}
